%% file: root.tex
\documentclass[letterpaper, 10 pt, conference]{ieeeconf}  

\usepackage{amsmath}

\IEEEoverridecommandlockouts                              

\usepackage{graphics} 
\usepackage{epsfig} 
\usepackage{mathptmx} 
\usepackage{times} 
\usepackage{amsmath} 
\usepackage{amssymb}  
\usepackage{adjustbox}

\usepackage{booktabs}

\title{\LARGE \bf
Modeling and Control of an Eel-Inspired Soft Robot for Design Optimization
}

\author{Zhangjingyi Jiang$^{1}$ and Mark Campbell$^{1}$
\thanks{$^{1}$Zhangjingyi Jiang, Mark Campbell are with the Department of Mechanical and Aerospace Engineering, Cornell University
        {\tt\small zj224@cornell.edu, mc288@cornell.edu}}%
}

\begin{document}

\maketitle
\thispagestyle{empty}
\pagestyle{empty}

\input{sections/abstract}

\input{sections/introduction}


\input{sections/model}

\input{sections/control}

\input{sections/ResultsAndDiscussion}



\input{sections/futurework}

\addtolength{\textheight}{-12cm}   

\bibliographystyle{IEEEtran}
\bibliography{main}

\end{document}

%% file: sections/abstract.tex
\begin{abstract}
Anguilliform locomotion is a highly efficient swimming mode; the advent of new materials for soft robots enables the development of an eel-inspired soft robot. This paper presents a simulation model of an eel-inspired soft robot designed for anguilliform swimming. This model can aid in design optimization and the development of model-based estimation, reasoning, and control systems. A Finite Element Method (FEM) model of an elastic rod is used to capture the soft materials of the robotic fish, which makes it particularly amenable to variation over time as the material properties change. The material model is coupled with a hydrodynamic force model to simulate the behavior of a soft, elongated robot in water. The model is used to demonstrate the effectiveness of the proposed control approaches in achieving desired swimming behaviors. It also provides insights into design decisions, including the robustness of different system configurations and the impact of material degradation and failure. The results show that slightly asymmetric designs are advantageous, offering comparable swimming velocities but greater maneuverability. This model can be used to guide future robotic design decisions aimed at optimizing performance for specific tasks.

\end{abstract}

%% file: sections/introduction.tex
\section{INTRODUCTION}
Anguilliform locomotion, the undulatory swimming style used by eels and other elongated vertebrates~\cite{undalatoryLocomotion}, is a highly efficient swimming mechanism~\cite{c4,c5,c6,c7}. Compared to trout, eels use four to six times less energy when swimming at the same speed~\cite{c4,c5,c6}. Most eel-like robots primarily focus on building a prototype and enabling basic swimming motion rather than optimizing design, improving swimming efficiency, or enhancing goal-seeking capabilities~\cite{c2,c8,selfOrganizedSwimming,c10}. 

The beauty of bio-inspired robotics lies in the ability to adapt designs for specific applications. A simulation platform that captures the physical characteristics of the soft robot and basic hydrodynamics enables the optimization of design parameters to meet key system goals, such as cruising and maneuvering, before building a physical prototype. Furthermore, such a model can be used to develop model-based algorithms for estimation, reasoning, and control.

\begin{figure}[thpb]
  \centering
  \includegraphics[scale=0.13]{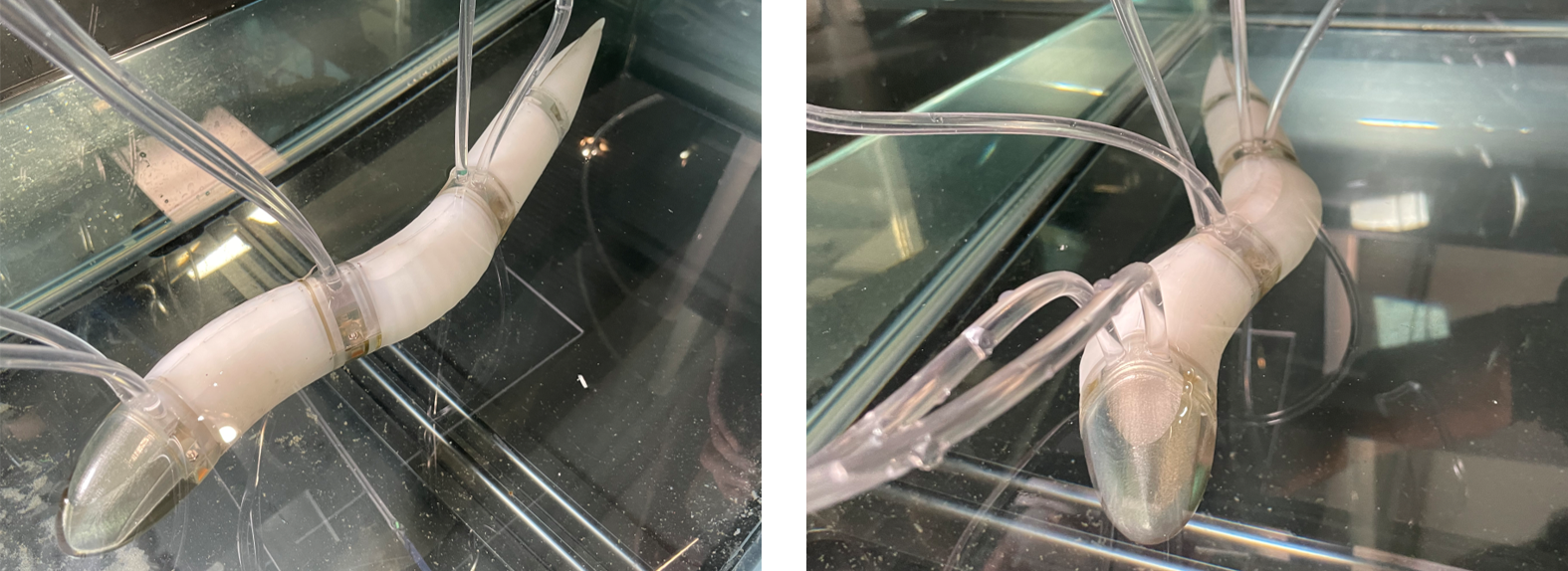}
  \caption{Physical fish robot from UCSD Bioinspired Robotics and Design Lab~\cite{c21}}
  \label{physicalFish}
\end{figure}

Autonomous underwater vehicles (AUVs) have many potential applications~\cite{c15}, one of which is approaching, finding, and sensing new, unexplored areas. A soft eel-like robot with autonomous anguilliform swimming capabilities could enable many impactful applications such as undersea infrastructure intelligent maintenance~\cite{c10}, deep sea exploration~\cite{c12}, and under-ice data collection~\cite{c13}. An example of such is the modular eel-inspired soft robot (Figure \ref{physicalFish}) from the Bio-inspired Robotics and Design Lab at the University of California San Diego (UCSD)~\cite{c21}. Many of these applications are challenging due to the harsh environmental conditions and communication difficulties~\cite{communicationDifficulties}. The robustness of autonomous robots continuing to operate in these conditions, even when material deterioration, damage, and failure occur, is essential in these deployments because it is extremely difficult and expensive, if not impossible, to locate and retrieve the robots. Optimizing system parameters (either offline or online) to facilitate sustained, efficient swimming that is resilient to damage and deterioration that may occur during operation will extend mission durations and support the robot's return to base for necessary repairs and maintenance.

%% file: sections/model.tex
\section{MODEL}

\begin{figure}[htbp]
  \centering
  \includegraphics[scale=0.5]{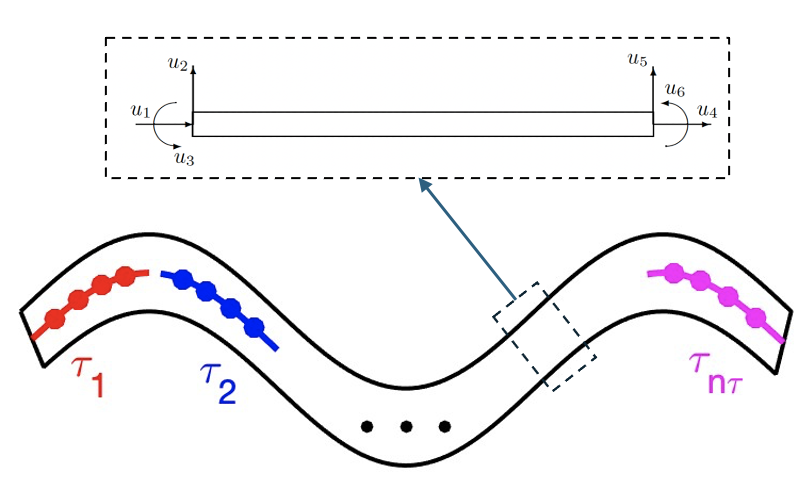}
  \caption{FEM model of fish robot with $n_\tau$ torque inputs. }
  \label{fishfig1}
\end{figure}

A Finite Element Method (FEM) model of a free-floating elastic rod (Figure \ref{fishfig1}) was used to simulate the internal forces and dynamics of the robotic fish. A simplified hydrodynamic force model was used to simulate the external fluid forces acting on the robot. The fish robot is split into $n_\tau$ sections, with each $\tau$ acting on every element within that section.
In this paper, "node" represents a finite element, and "actuator" represents a collection of nodes with the same torque input. In Figure \ref{fishfig1}, there are $n_\tau$ actuators, each with 4 nodes.

\subsection{FEM Model of the Soft Robot}

A 2-dimensional model of an elastic rod with free-free ends is generated using Finite Element Method (FEM). The rod is separated into $n_e$ number of elements. A set number of torque input actuators $n_{\tau}$ are distributed across the rod, simulating $n_{\tau}+1$ links. Each node has 3 degrees of freedom: $\mathbf{q_i} = \begin{bmatrix} x_i \ y_i \ \theta_i \end{bmatrix}'$.  Stacking $n_e$ elements together, the model is as follows:

\begin{equation}
M \mathbf{\ddot{q}}(t) + K \mathbf{q}(t) = \mathbf{Q}(t)
\end{equation}

where $M$ and $K$ are the global mass and stiffness matrices respectively. $\mathbf{Q}$ is a vector of global forces, and $\mathbf{q}$ is a vector of $n=3n_e$ global degrees of freedom.

\begin{equation}
\mathbf{Q}(t) = \beta_{w}\mathbf{w} + \beta_{u}\mathbf{u}
\end{equation}

where the input forces are expressed in terms of disturbances $\mathbf{w}$ and control inputs $\mathbf{u}$, $\beta_{w}$ is the disturbance input influence coefficient, and  $\beta_{u}$ is the control input influence coefficient.

The global mass and stiffness matrices $M$ and $K$ can be derived from the mass and stiffness matrices for individual beam finite elements: 
\begin{equation}
m_i = \frac{\rho_r A_c l}{420}
\begin{bmatrix}
140 & 0 & 0 & 70 & 0 & 0 \\
0 & 156 & 22l & 0 & 54 & -13l \\
0 & 22l & 4l^2 & 0 & 13l & -3l^2 \\
70 & 0 & 0 & 140 & 0 & 0 \\
0 & 54 & 13l & 0 & 156 & -22l \\
0 & -13l & -3l^2 & 0 & -22l & 4l^2
\end{bmatrix}
\end{equation}

\begin{equation}
k_i = \frac{E I}{l^3}
\begin{bmatrix}
\frac{l^2 A_c}{I} & 0 & 0 & -\frac{l^2 A_c}{I} & 0 & 0 \\
0 & 12 & 6l & 0 & -12 & 6l \\
0 & 6l & 4l^2 & 0 & -6l & 2l^2 \\
-\frac{l^2 A_c}{I} & 0 & 0 & \frac{l^2 A_c}{I} & 0 & 0 \\
0 & -12 & -6l & 0 & 12 & -6l \\
0 & 6l & 2l^2 & 0 & -6l & 4l^2
\end{bmatrix}
\end{equation}

where $\rho_r$ is the robot (beam) density, $A_c$ is the cross sectional area, $l$ is the length of beam segment, $E$ is the Young's Modulus of the beam material, $I$ is the area moment of inertia.

The variables used are shown in Table \ref{FEMModelParameters}. These values were inspired by the modular eel-inspired soft robot (Figure \ref{physicalFish})~\cite{c21}, with modifications made to reflect a larger fish robot with more active actuators.

\begin{table}[hb]
    \centering
    \caption{FEM Model Parameters}
    \label{FEMModelParameters}
    \begin{tabular}{@{}cc@{}}
        \toprule
        Variables & Values \\
        \midrule
        Number of Elements $n_e$ & $20$ \\
        Number of Torque Inputs (Actuators) $n_\tau$ & $5$ \\ 
        Material Density $\rho_r \ (kg \cdot m^{-3})$ & $1200$ \\ 
        Young's Modulus of Material $E \ (N \cdot m^{-2})$ & $1 \ x \ 10^6$ \\ 
        Rod Length $L \ (m)$ & $1$ \\ 
        Rod Cross-Sectional Area $A_C \ (m^2)$ & $7.85 \ x \ 10^{-3}$ \\ 
        Surface Area of an Element $s \ (m^2)$ & $2 \pi r l/n_e$ \\
        Torque Input Locations $\ell^\tau_{1:n\tau} \ (m)$ & $[\frac{1}{6} \ \frac{1}{3} \ \frac{1}{2} \ \frac{2}{3} \ \frac{5}{6}]$ \\ 
        \bottomrule
    \end{tabular}
\end{table}

A key attribute of this model is that each element is parameterized. This allows the parameters of each element to be independent, thus being capable of capturing the difference in material properties along the soft robot in addition to the change in material properties over time. 

\subsection{Hydrodynamic Force Model} 

The drag force for an object moving in a fluid can be expressed as follows:

\begin{equation}
F_{D} = \frac{1}{2} C_{D} \rho_f s v^{2}
\end{equation}

where $C_{D}$ is the drag coefficient, $\rho_f$ is the fluid density, $s$ is surface area, $v$ is object velocity~\cite{c3}. 

Drag forces are assumed perpendicular to the robot surface because shear forces are insignificant at relatively high Reynolds numbers (Re) where many fish operate ~\cite{ignoreShear1,ignoreShear2}. The hydrodynamic force acting on the side of the robot for each finite element can be expressed as:

\begin{equation}
\mathbf{F_{Dside}} = -\frac{1}{2} C_D \rho_f s \cdot \mathrm{sign}(\mathbf{V} \cdot \mathbf{e}_{\mathrm{\theta}}) (\mathbf{V} \cdot \mathbf{e}_{\mathrm{\theta}})^2 \cdot \mathbf{e}_{\mathrm{\theta}}
\end{equation}

where $\mathbf{e}_{\theta}$ is the transverse unit vector indicating the direction perpendicular to the finite element, and $\mathbf{V}$ is the velocity vector of the finite element, calculated as:

\begin{equation}
\mathbf{V} = \begin{bmatrix} \dot{x} \\ \dot{y} \end{bmatrix} = \begin{bmatrix} 1 \ 0 \ 0 \\ 0 \ 1 \ 0 \end{bmatrix} \dot{q}
\end{equation}

The hydrodynamic drag force acting on the tip of the robot head can be expressed as:

\begin{equation}
\mathbf{F_{Dhead}} = -\frac{1}{2} C_D \cdot \rho_f s_{eff} \cdot \mathrm{sign}(\mathbf{V_{head}} \cdot \mathbf{e}_r) (\mathbf{V_{head}} \cdot \mathbf{e}_r)^2 \cdot \mathbf{e}_r
\end{equation}

where $\mathbf{e}_r$ is the radial unit vector indicating the orientation of the robot head, $\mathbf{V_{head}}$ is the velocity vector of the head, and $\mathbf{s_{eff}}$ is the effective surface area of the head, calculated as $\frac{1}{4}$ the cross-sectional area of the head. 

Therefore, the total drag force can be expressed as:

\begin{equation}
\mathbf{F_{D}} = \mathbf{F_{Dhead}} + \mathbf{F_{Dside}}
\end{equation}

The variables used are shown in Table \ref{HydroParameters}. The drag coefficient $C_D$ was chosen based on previous measurements done on another eel-inspired soft robot~\cite{c2}. The fluid density $\rho_f$ was chosen to be that of water because initial experiments for a physical fish robot will be done in water. 

\begin{table}[hb]
    \centering
    \caption{Hydrodynamic Force Parameters}
    \label{HydroParameters}
    \begin{tabular}{@{}cc@{}}
        \toprule
        Variables & Values \\
        \midrule
        Drag Coefficient $C_D$ & $0.167$~\cite{c2} \\ 
        Fluid Density $\rho_f \ (kg/m^3)$ & $10^3$ \\ 
        \bottomrule
    \end{tabular}
\end{table}

%% file: sections/control.tex
\section{CONTROLS}

Like other elongated anguilliform swimmers, Eels generate motion by undulating their bodies with traveling waves~\cite{c19,c20}. Previous works on anguilliform swimming robots have utilized many forms of control, including pulse signals with shift phase to control actuator valves~\cite{c2}, predetermined controls for different gaits (forward, turning, sideways, spinning)~\cite{c16}, and a receding horizon controller~\cite{c17}. We chose to use an open-loop traveling sinusoidal wave with bias as the torque input because it produces a natural swimming motion while also being simple and efficient, allowing for optimization for swimming inputs.

\subsection{Linear Velocity Control}

The torque input on each joint actuator for linear swimming can be expressed as:

\begin{equation}
\tau_j = A_{j} \sin(\omega t + \phi_j)
\end{equation}
where $j$ is the joint index, $\tau_j$ is the torque input on the $j^{th}$ joint, $A_{j}$ is the amplitude for the $j^{th}$ joint, $\omega$ is the oscillation frequency, $t$ is the current time, $\phi_j$ is the phase shift for the $j^{th}$ joint. The input vector is then:

\begin{equation}
u = [\tau_1 \ \tau_2 \ \ldots \ \tau_{n\tau}]
\end{equation}

The set of $n_\tau$ amplitude ratios $A_{1:n\tau} = [A_1:A_2:\ldots:A_{n\tau}]$ and $\omega$ can be optimized to have the highest steady-swimming linear velocity for the current shape design. The magnitude of the linear velocity can be controlled by multiplying the optimized $A_{1:n\tau}$ by a constant $C_A \in [0, 1]$ or by multiplying the optimized $\omega$ by a constant $C_{\omega} \in [0, 1]$. Backward swimming can be achieved by reversing $\phi_j$. The modified linear velocity control torque can be expressed as:

\begin{equation}
u = [C_A \cdot A_{1:n\tau}] \sin(C_\omega \cdot \omega t + \phi_{1:n\tau})
\end{equation}

\subsection{Angular Velocity Control}

Turning can be achieved by adding a torque offset to the linear velocity control. The torque input for turning can be expressed as:

\begin{equation}
u = [C_A \cdot A_{1:n\tau}] \sin(C_\omega \cdot \omega t + \phi_{1:n\tau}) + \tau^O_{1:n\tau}
\end{equation}

The torque offset ratio $\tau^O_{1:n\tau} = [\tau^O_{1}:\tau^O_{2}:\ldots:\tau^O_{n\tau}]$ can be optimized to have the highest steady-swimming angular velocity for the current shape design. The magnitude of angular velocity and turning radius can be controlled by multiplying the optimized $\tau^O_{1:n\tau}$ by a constant $C^O_{\tau} \in [-1, 1]$. Turning direction is controlled by the direction of the torque offset. The modified angular velocity control torque can be expressed as:

\begin{equation}
u = [C_A \cdot A_{1:n\tau}] \sin(C_\omega \cdot \omega t + \phi_{1:n\tau}) + C^O_{\tau} \cdot \tau^O_{1:n\tau}
\end{equation}

The variables used are shown in Table \ref{ControlParameters}. The amplitude ratio $A_{1:n\tau}$ was found by sweeping combinations of amplitudes $A_j \in [0,5]$ and choosing $A_{1:n\tau}$ with the highest steady swimming linear velocity. The torque offset ratio $\tau^O_{1:n\tau}$ was found by sweeping combinations of amplitudes $\tau^O_j \in [0,5]$ and choosing $\tau^O_{1:n\tau}$ with the highest steady swimming angular velocity. The oscillation frequency $\omega$ was chosen based on the predicted limitations of the physical robot actuators. 

\begin{table}[hb]
    \centering
    \caption{Control Parameters}
    \label{ControlParameters}
    \begin{tabular}{@{}cc@{}}
        \toprule
        Variables & Values \\
        \midrule
        Amplitude Ratio $A_{1:n\tau}$  & $[20 \ 20 \ 16 \ 4 \ 4]$ \\
        Oscillation Frequency $\omega$*$\ (Hz)$  & $1.5$ \\
        Phase Shift $\phi \ (rads)$ & $2 \pi/n_\tau \approx 1.2566$ \\
        Joint Index $j_{1:n\tau}$ & $[1 \ 2 \ 3 \ 4 \ 5]$ \\
        Torque Offset Ratio $\tau^O_{1:n\tau}$  & $[3.5 \ 5.5 \ 6.5 \ 6.5 \ 3.5]$ \\
        \bottomrule
        \multicolumn{2}{l}{*all actuators are oscillating at the same frequency} \\
    \end{tabular}
\end{table}

%% file: sections/ResultsAndDiscussion.tex
\section{RESULTS AND DISCUSSION}
In this paper, the model was used to demonstrate the simulated fish robot undergoing linear swimming, turning, and accelerating from and decelerating to rest. The model was then used to study three key design questions, the robustness of a symmetrical design to torque input degradation/failure, the performance of symmetric vs. asymmetric designs, and the robustness of asymmetrical designs to actuator degradation/failure. Steady swimming linear velocity and turning radius were used as a measure of robustness, with lower turning radius indicating the ability to make tighter turns.

In sections B, C, D, all of the velocities and turning radii mentioned are optimal values. Thus the robot fish can be made to swim at any velocity lower than said values and any turning radii higher than said values by adjusting the $C_A,C_\omega$, and $C^O_\tau$ values.

The simulations were run on a 2.4 GHz 8-Core i9 processor. A current iteration of this simulator can run at over 3300 Hz (time-steps per second). 

\subsection{Demonstration of Controls}

\begin{figure}[htbp]
  \centering
  \includegraphics[scale=0.5,trim=1cm 3cm 1.5cm 3cm,clip]{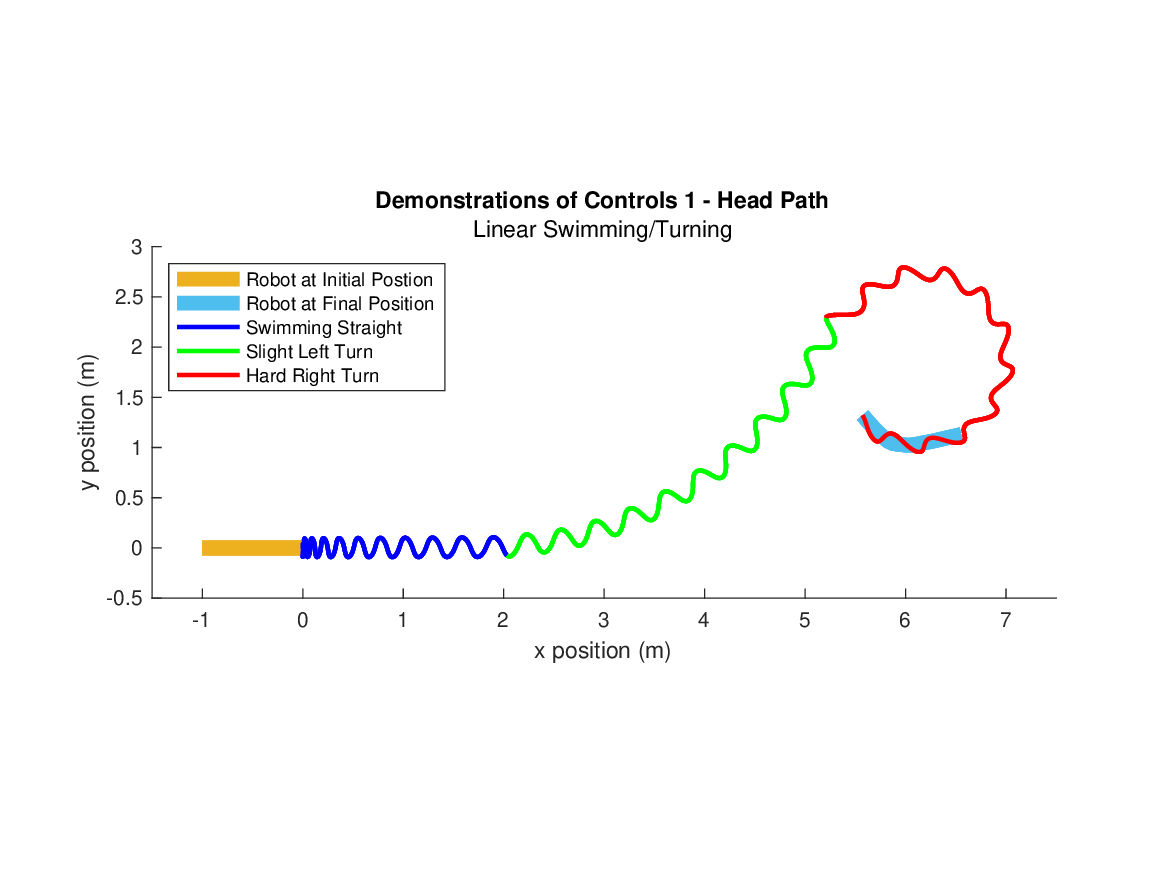}
  \caption{Demonstration of the robot swimming straight (blue), slight turning left (green), and hard turning right (red) for $t = 15s$ each}
  \label{demo1}
\end{figure}

Three sets of controls (Table \ref{ControlSettings1}) are enacted on the robot to demonstrate swimming straight, slight turning, and hard turning capabilities on the simulated robot (Figure \ref{demo1}). The results show the robot accelerating from rest and varying levels of turning. Adjusting the control parameters is intuitive (increasing $C_A,C_\omega$ increases velocity, and increasing the magnitude of $C^O_\tau$ decreases the turning radius for tighter turns), and there is a smooth transition between different sets of controls.

\begin{table}[hb]
    \centering
    \caption{Control Parameters - Demo 1}
    \label{ControlSettings1}
    \begin{tabular}{@{}ccccc@{}}
        \toprule
        Swimming & $C_A$ & $C_\omega$ & $C^O_\tau$ & Time $\Delta t$ (s) \\
        \midrule
        Swimming Straight & $1$ & $1$ & $0$ & $7$\\
        Slight Left Turn & $1$ & $1$ & $-0.2$ & $7$\\
        Hard Right Turn & $0.75$ & $1$ & $1$ & $7$\\
        \bottomrule
    \end{tabular}
\end{table}

\begin{figure}[htbp]
  \centering
  \includegraphics[scale=0.5,trim=1cm 0.5cm 1.5cm 0.5cm,clip]{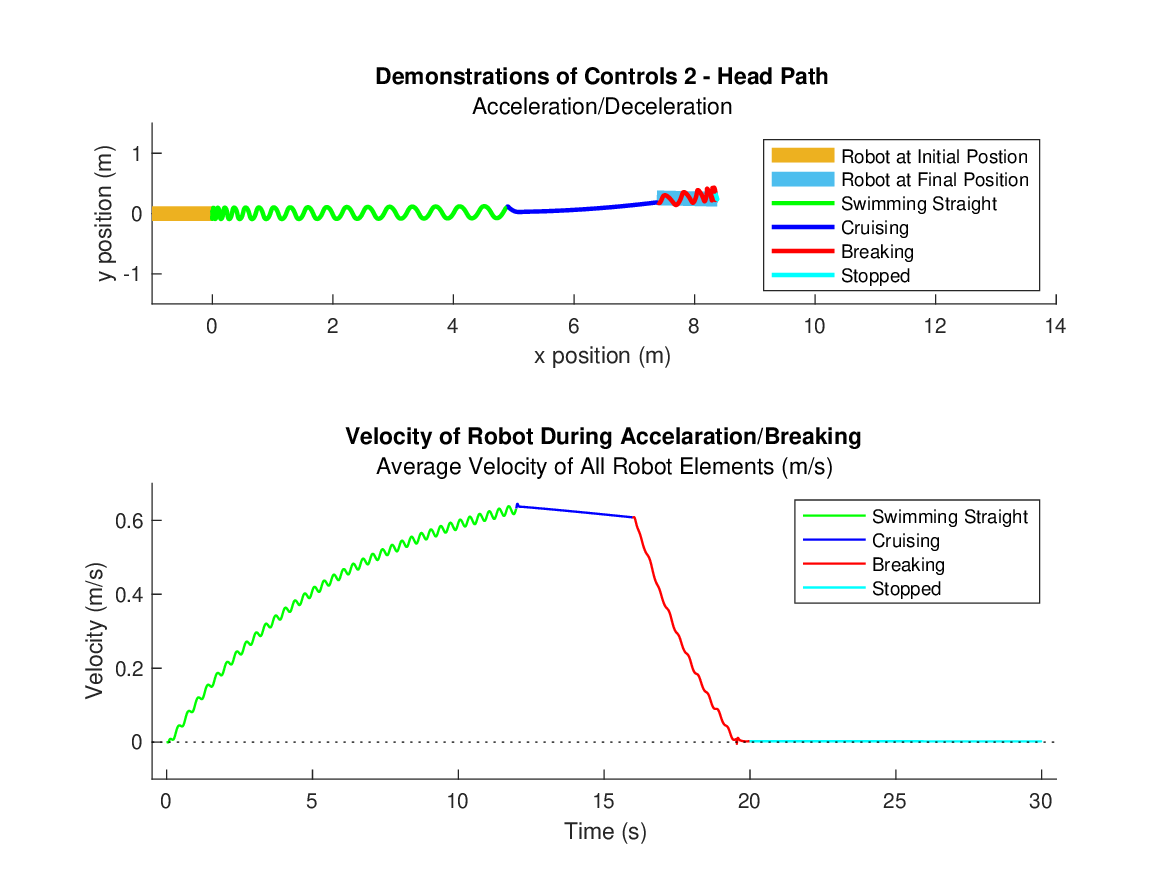}
  \caption{Demonstration of the robot accelerating (green), cruising (blue), breaking (red), and stopped (cyan) for $12s, 4s, 4s, 10s$ respectively}
  \label{demo2}
\end{figure}

Four sets of controls (Table \ref{ControlSettings2}) are enacted on the robot to demonstrate acceleration from and breaking to rest on the simulated robot (Figure \ref{demo2}). The results show that the robot can accelerate from rest and cruise using the controls proposed above, and the robot can successfully decelerate to rest by reversing the phase shift and flipping the amplitude ratio $A_{1:n\tau}$.

\begin{table}[hb]
    \centering
    \caption{Control Parameters Demo 2}
    \label{ControlSettings2}
    \begin{tabular}{@{}ccccc@{}}
        \toprule
        Swimming & $C_A$ & $C_\omega$ & $C^O_\tau$ & Time $\Delta t$ (s) \\
        \midrule
        Accelerate & $1$ & $1$ & $0$ & $12$\\
        Cruise & $0$ & $0$ & $0$ & $4$\\
        Break* & $1$ & $1$ & $0$ & $3.55$\\
        Stopped & $0$ & $0$ & $0$ & $10.45$\\
        \bottomrule
        \multicolumn{5}{l}{*When breaking, $\phi$ becomes negative and $A_{1:n\tau}$ is flipped} \\
    \end{tabular}
\end{table}

\subsection{Robustness to Actuator Degradation/Failure}

AUVs face unique challenges compared to other autonomous robots, including damage from marine life swimming into or latching onto the robot and the degradation of components over time. Communication difficulties and the harsh deployment environments make retrieval of the robot difficult. Therefore, the ability of the robot to swim back to a base station, even with damage or degradation, is an important design requirement. 

The model can be used to evaluate the robustness of an individual design to damage and degradation by changing the actuator torque input to reflect actuator degradation, modifying the design parameters to reflect changes of mass and shape due to interactions with the environment, or changing the control law to reflect breakdowns in communication between actuators.

\begin{figure}[htbp]
  \centering
  \includegraphics[scale=0.5,trim=1cm 0cm 1.5cm 0cm,clip]{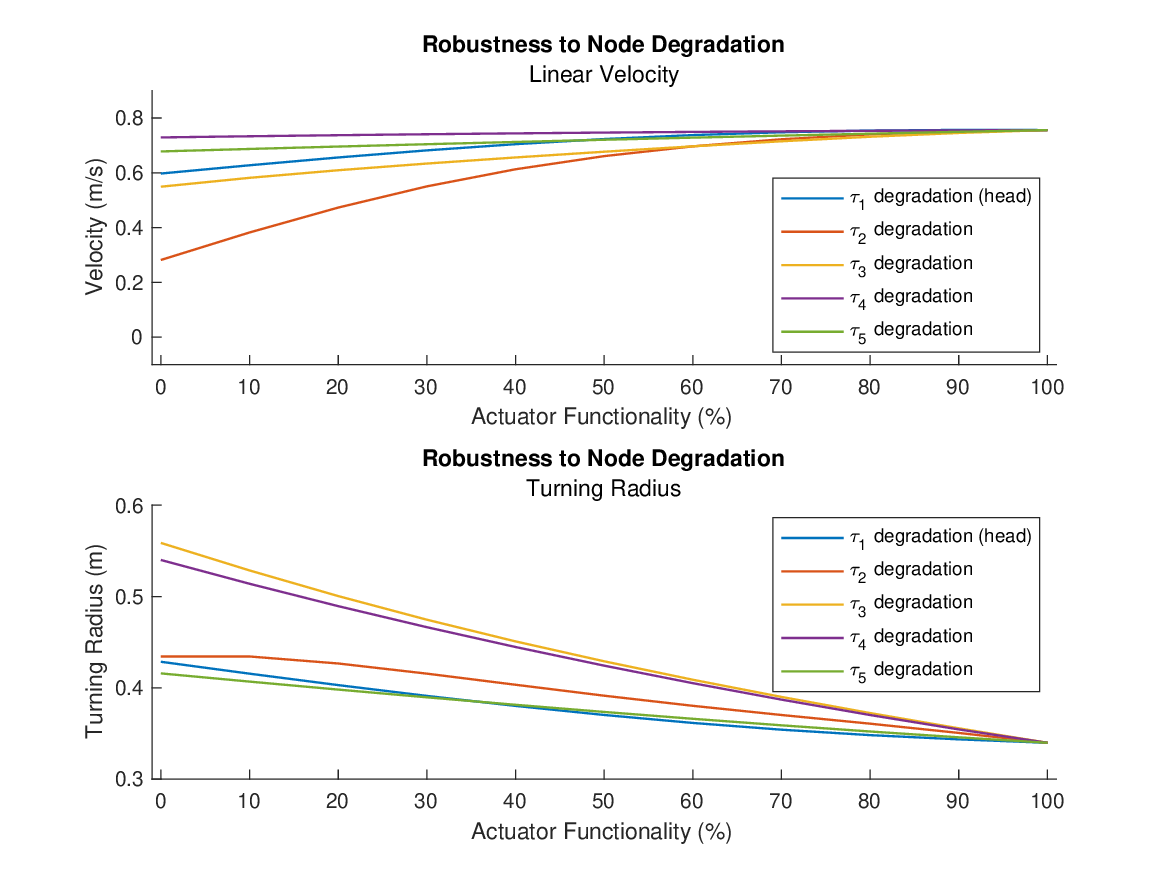}
  \caption{Steady swimming linear velocities and turning radii when torque inputs are operating below maximum}
  \label{rob1}
\end{figure}

In this study, we chose to investigate the robustness of an eel-like robotic fish to actuator degradation and failure as a demonstration of model capabilities. Actuator degradation was measured by actuator functionality, where $0\%$ represents the actuator fully turned off. The results (Figure \ref{rob1}) show the steady-swimming linear velocities were more affected by the degradation of torque actuators in the front of the fish ($\tau_{1:3}$), especially actuator 2 ($\tau_{2}$), which is consistent with findings from other studies \cite{c23}. The results also show turning radii were more affected by torque actuators towards the back of the fish ($\tau_{3:4}$)

\subsection{Performance of Symmetric vs. Asymmetric Designs}

To demonstrate how the model can be used to test different shape designs, we varied the fish robot head and tail cross-sectional areas to produce asymmetrical designs while keeping similar surface area and volume (Table \ref{DesignParameters}). Steady swimming linear velocities and turning radii were used as a measure of performance. Turning radii were calculated by fitting the steady-state swimming path to a circle and calculating the radius. 

\begin{table}[hb]
    \centering
    \caption{Design Parameters}
    \label{DesignParameters}
    \begin{tabular}{@{}ccccc@{}}
        \toprule
        $A_H$ ($m^2$) & $A_T$ ($m^2$) & $A_H/A_T$ & $SA$ ($m^2$) & V ($10^3 m^3$) \\
        \midrule
        0.30$\pi$ & 0.10$\pi$ & 1.5/0.5 & 0.31 & 8.5 \\
        0.29$\pi$ & 0.11$\pi$ & 1.45/0.45 & 0.31 & 8.4 \\
        0.28$\pi$ & 0.12$\pi$ & 1.4/0.6 & 0.31 & 8.3 \\
        0.27$\pi$ & 0.13$\pi$ & 1.35/0.65 & 0.31 & 8.2 \\
        0.26$\pi$ & 0.14$\pi$ & 1.3/0.7 & 0.31 & 8.1 \\
        0.25$\pi$ & 0.15$\pi$ & 1.25/0.75 & 0.31 & 8.0 \\
        0.24$\pi$ & 0.16$\pi$ & 1.2/0.8 & 0.31 & 8.0 \\
        0.23$\pi$ & 0.17$\pi$ & 1.15/0.85 & 0.31 & 8.9 \\
        0.22$\pi$ & 0.18$\pi$ & 1.1/0.9 & 0.31 & 7.9 \\
        0.21$\pi$ & 0.19$\pi$ & 1.05/0.95 & 0.31 & 7.9 \\
        0.20$\pi$ & 0.20$\pi$ & 1/1 & 0.31 & 7.9 \\
        \bottomrule
        \multicolumn{5}{l}{$A_H$: Cross-sectional area of the head} \\
        \multicolumn{5}{l}{$A_T$: Cross-sectional area of the tail} \\
        \multicolumn{5}{l}{$A_H/A_T$: Head-to-tail ratio} \\
        \multicolumn{5}{l}{$SA$: Surface Area} \\
    \end{tabular}
\end{table}

\begin{figure}[htbp]
  \centering
  \includegraphics[scale=0.43,trim=1cm 3cm 0cm 0cm,clip]{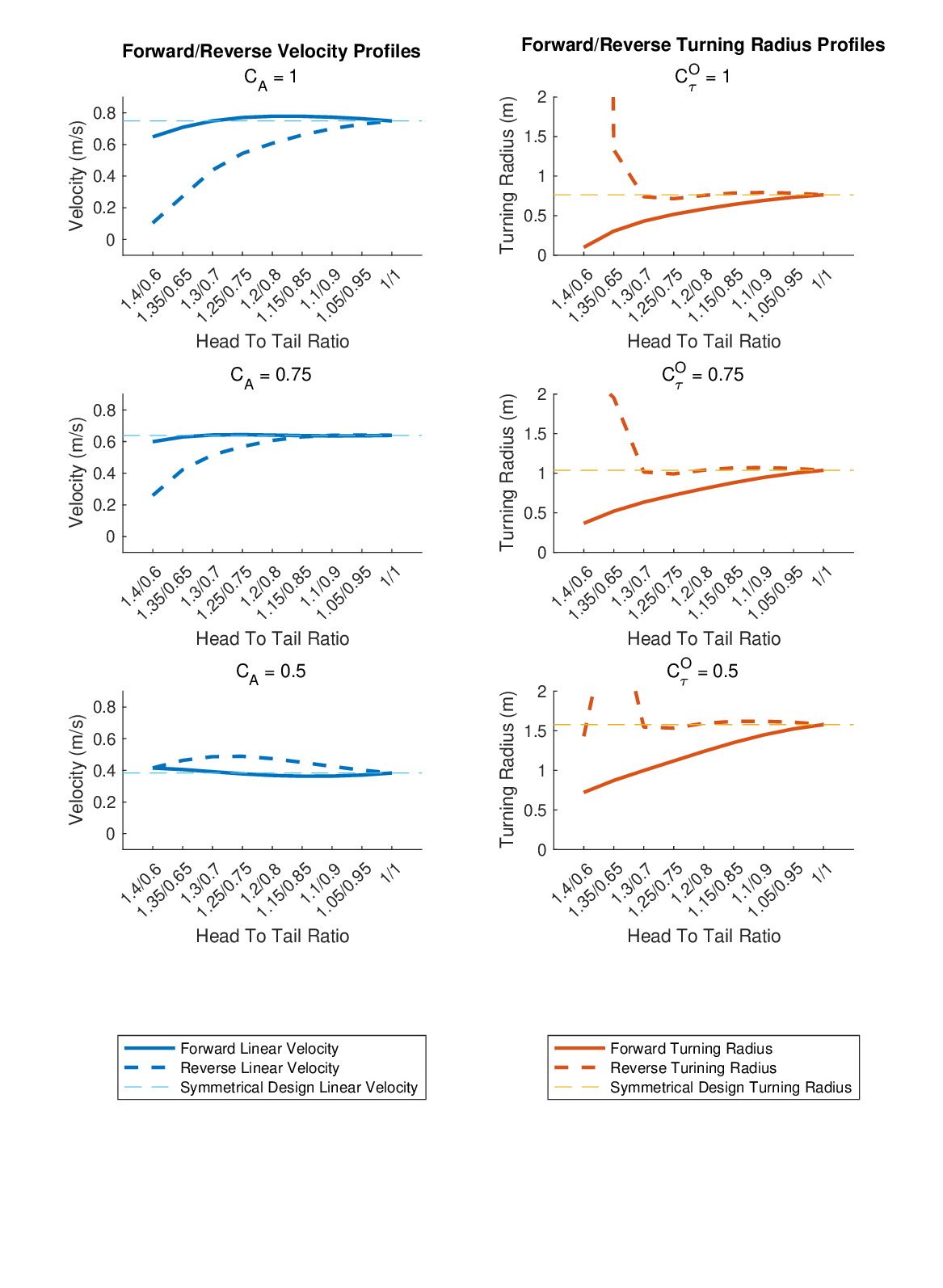}
  \caption{A comparison of steady swimming linear velocities for symmetric and asymmetric robot designs at $C_A = 1$,$C_A = 0.75$, and $C_A = 0.5$ (left column). A comparison of turning radii for symmetric and asymmetric robot designs at $C^O_\tau = 1$,$C^O_\tau = 0.75$, and $C^O_\tau = 0.5$ (right column).}
  \label{symLin}
\end{figure}

The results (Figure \ref{symLin}) show for $C_A = 1$, slightly asymmetric designs had comparable forward swimming velocity as the symmetric design, but lower reverse swimming velocity. For $C_A = 0.75$, slightly asymmetric designs had similar forward and reverse velocities as symmetric designs. For $C_A = 0.5$, asymmetric designs had higher reverse velocities than symmetric designs. As for turning radii, slightly asymmetric designs had much lower forward turning radii and similar reverse turning radii when compared to symmetric designs. When head to tail ratio drops before 1.3/0.7, the reverse turning radii greatly increases. 

A fish robot deployed in the ocean will swim at below-maximum velocity most of the time. In addition, high maneuverability is needed to swim around obstacles. Therefore, a slightly asymmetric design (1.25/0.75) is recommended.

\subsection{Robustness of Asymmetrical Designs to Actuator Degradation/Failure}

\begin{figure}[htbp]
  \centering
  \includegraphics[scale=0.45,trim=1cm 2.5cm 1cm 0cm,clip]{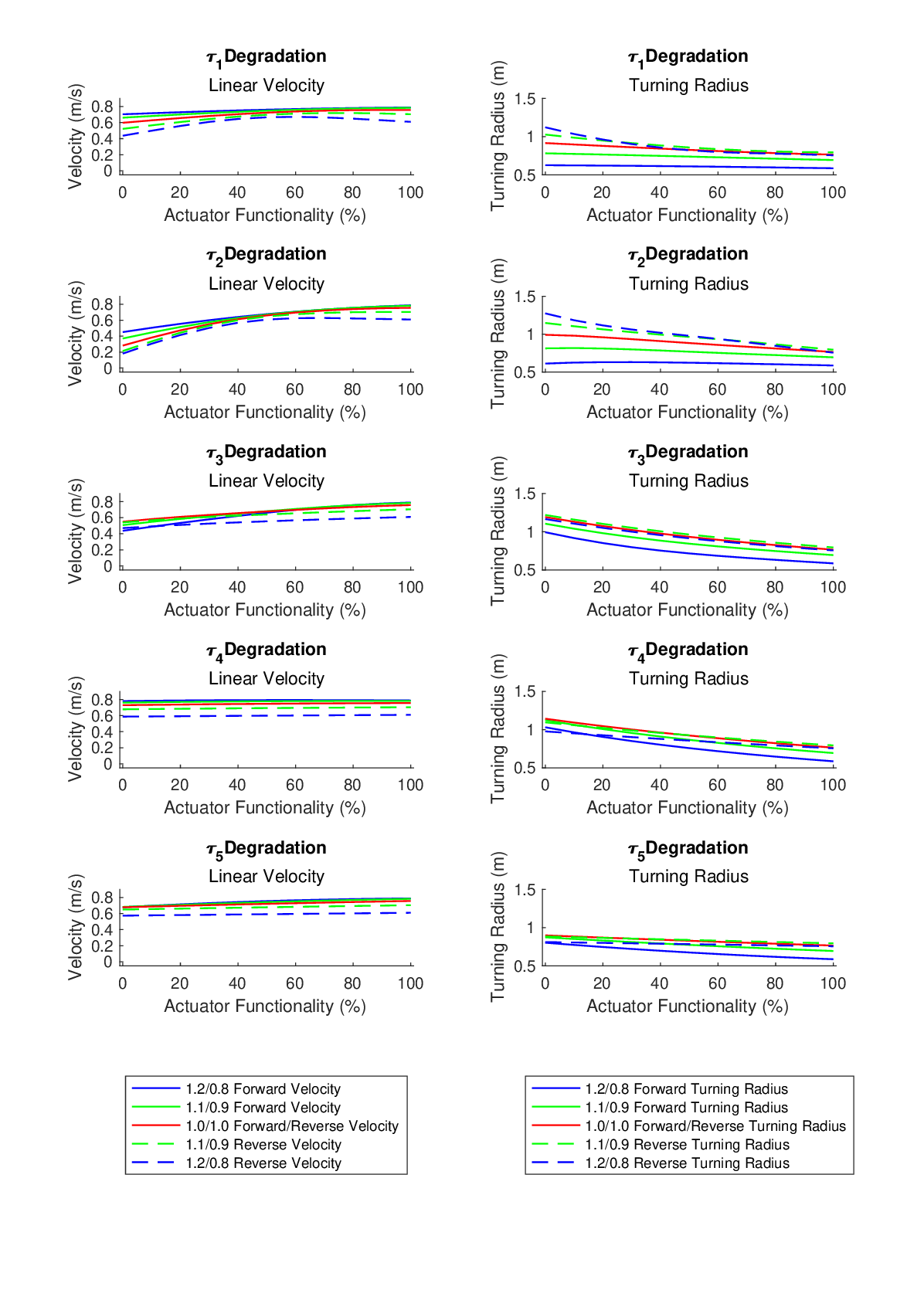}
  \caption{Linear velocities and turning radii for design 1 (rod with length l = 1m rod with $A_H=0.30\pi m^2$  and $A_T=0.10\pi m^2$ - blue), design 2 (l = 1m, $A_H=0.25\pi m^2$  and $A_T=0.15\pi m^2$  - green), and design 3 (l = 1m, $A_H=0.20\pi m^2$  and $A_T=0.20\pi m^2$  - red) for forward (solid line) and reverse (dotted line) steady swimming.}
  \label{nonsym}
\end{figure}

Swimming behaviors were measured for various actuator functionalities to test the robustness of asymmetrical designs to actuator degradations and failures. The results (Figure \ref{nonsym}) show actuator degradations towards the head have a larger impact on linear velocity for all three designs. It also shows $\tau_{3}$ and $\tau_{4}$ degradation have a larger impact on turning radii on both designs, whereas $\tau_{1}$ and $\tau_{2}$ mostly impacts reverse turning radii for asymmetric designs. This is consistent with results from previous sections.

%% file: sections/futurework.tex
\section{CONCLUSIONS AND FUTURE WORK}
This paper developed a simulation model for design optimization and controller testing of a fish robot and proposed a control method for anguilliform swimming. Results show the control law can successfully be used to swim straight, complete soft and hard turns, and accelerate from and decelerate to rest. A comparison of steady swimming linear velocities and turning radii when undergoing actuator degradation shows the actuators towards the front of the fish body have a higher impact on linear swimming, and actuators towards the back have a higher impact on turning radii when degraded. A comparison of symmetric vs asymmetric robot designs shows asymmetric designs provide comparable swimming velocities but greater maneuverability. 

The results demonstrate the model's capability to assess the performance of different designs and their robustness to actuator degradation and failure. By manipulating the actuator torque input, adjusting the design parameters to accommodate additional or reduced mass, or modifying the control law to address breakdowns in communication between actuators, the model can be used to offer insights into the resilience of a variety of designs under diverse conditions.

Some notable advantages of the model include its modular nature and efficient simulation capabilities (over 3300 Hz), enabling seamless integration with different robot models tailored to specific design specifications while facilitating rapid iteration and evaluation of design iterations. Furthermore, the model facilitates the evaluation of controller designs, allowing for testing in different scenarios and environments. The model can also be used to enable fault detection and recovery in addition to design optimization-based controllers such as model predictive control (MPC) controllers. 

The model can optimize and fine-tune the design using cost functions that encompass parameters such as steady swimming behaviors, size and mass considerations, power consumption, and performance in completing tasks such as navigating digital obstacle courses. After predefined design goals are met, filters can be implemented to refine the optimization process. 

Overall, the proposed model offers a comprehensive framework for assessing and optimizing designs, encompassing various factors such as robustness, controller performance, modularity, cost considerations, and simulation speed, thereby supporting informed decision-making in the design and development of robotic systems.